\documentclass[runningheads]{llncs}

\usepackage[mobile]{eccv}

\makeatletter
\let\llncssubparagraph\subparagraph
\let\subparagraph\paragraph
\usepackage{titlesec}
\titleformat{\paragraph}[runin]{\normalfont\normalsize\bfseries}{\theparagraph}{1em}{}
\let\subparagraph\llncssubparagraph
\makeatother

\usepackage{eccvabbrv}

\usepackage{graphicx}
\usepackage{booktabs}

\usepackage[accsupp]{axessibility}  % Improves PDF readability for those with disabilities.

\usepackage{pifont}
\usepackage{multirow}
\usepackage[table]{xcolor}
\usepackage{listings}
\usepackage{marvosym}

\lstdefinestyle{promptbox}{
  backgroundcolor=\color{gray!12},
  basicstyle=\small\ttfamily,
  breaklines=true,
  breakatwhitespace=false,
  postbreak=\mbox{\textcolor{gray}{$\hookrightarrow$}\space},
  frame=none,
  xleftmargin=4pt,
  xrightmargin=4pt,
  aboveskip=6pt,
  belowskip=6pt,
  columns=fullflexible,
  keepspaces=true,
  showstringspaces=false,
}

\usepackage{hyperref}

\usepackage{orcidlink}

\titlespacing*{\section}{0pt}{10pt plus 2pt minus 2pt}{6pt plus 1pt minus 1pt}
\titlespacing*{\subsection}{0pt}{8pt plus 2pt minus 2pt}{6pt plus 1pt minus 1pt}
\titlespacing*{\subsubsection}{0pt}{6pt plus 2pt minus 1pt}{4pt plus 1pt minus 1pt}
\titlespacing*{\paragraph}{0pt}{5pt plus 1pt minus 1pt}{0.5em}

\titleformat{\subsubsection}[runin]{\normalfont\small\bfseries}{\thesubsubsection}{1em}{}

\begin{document}

\title{UENR-600K: A Large-Scale Physically Grounded Dataset for Nighttime Video Deraining} 
% Unlocking Generative Priors for Nighttime Video Deraining with a Large-Scale Physically Grounded Dataset

\titlerunning{UENR-600K}

% Second Author\inst{2,3}\orcidlink{1111-2222-3333-4444} \and
% Third Author\inst{3}\orcidlink{2222--3333-4444-5555}}


% Springer Heidelberg, Tiergartenstr.~17, 69121 Heidelberg, Germany

\author{Pei Yang\inst{1}\textsuperscript{*} \and
Hai Ci\inst{1}\textsuperscript{*} \and
Beibei Lin\inst{2} \and
Yiren Song\inst{1} \and
Mike Zheng Shou\inst{1}\textsuperscript{\Letter}}

\authorrunning{P. Yang et al.}

\institute{Show Lab, National University of Singapore \\
\email{yangpei@u.nus.edu, cihai03@gmail.com, yiren@u.nus.edu, mike.zheng.shou@gmail.com}
\and
National University of Singapore \\
\email{beibei.lin@u.nus.edu}\\[0.5ex]
\textsuperscript{*} Equal contribution. \qquad \textsuperscript{\Letter} Corresponding author.}

% \author{%
% \makebox[\textwidth][c]{%
% \parbox[t]{0.30\textwidth}{\centering
% Auther One\textsuperscript{*}\\
% Shared Institution, City, Country\\
% author1@shared.edu
% }
% \hfill
% \parbox[t]{0.30\textwidth}{\centering
% Author Two\textsuperscript{*}\\
% Shared Institution, City, Country\\
% author2@shared.edu
% }
% \hfill
% \parbox[t]{0.30\textwidth}{\centering
% Author Three\\
% Different Institution, City, Country\\
% author3@different.edu
% }
% }
% \\[1.2em]
% \makebox[\textwidth][c]{%
% \parbox[t]{0.30\textwidth}{\centering
% Author Four\\
% Shared Institution, City, Country\\
% author4@shared.edu
% }
% \hfill
% \parbox[t]{0.30\textwidth}{\centering
% Author Five\textsuperscript{\Letter}\\
% Shared Institution, City, Country\\
% author5@shared.edu
% }
% }
% }

% \authorrunning{A. One et al.}

% \institute{\textsuperscript{*} Equal contribution. \qquad \textsuperscript{\Letter} Corresponding author.}

\maketitle

\begin{center}
    \url{https://showlab.github.io/UENR-600K/}
\end{center}

\vspace{-10pt}

\begin{figure}
    \centering
    \includegraphics[width=\linewidth]{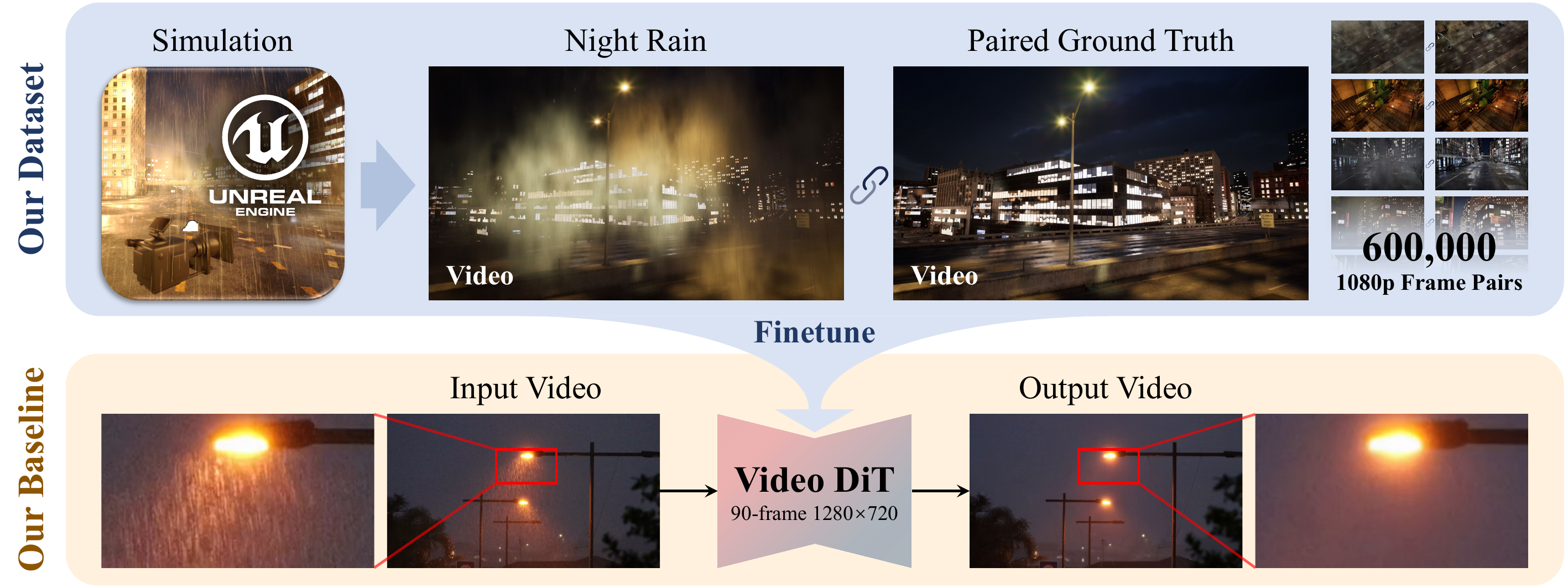}
    \caption{Paper Overview. \textbf{Top:} We use Unreal Engine 5 to simulate rain as 3D particles within virtual environments, producing 600,000 paired 1080p frames with physically grounded nighttime rain. Left: a rainy video frame; right: the paired ground truth. \textbf{Bottom:} We finetune a video Diffusion Transformer on our dataset for nighttime video deraining. Given a real nighttime rain video (left), our baseline removes rain and rain-induced fog near the streetlamp while preserving scene detail (right; see red crop).}
    \label{fig:teaser}
\end{figure}

\vspace{-20pt}

\begin{abstract}
    Nighttime video deraining is uniquely challenging because raindrops interact with artificial lighting. Unlike daytime white rain, nighttime rain takes on various colors and appears locally illuminated. Existing small-scale synthetic datasets rely on 2D rain overlays and fail to capture these physical properties, causing models to generalize poorly to real-world night rain. Meanwhile, capturing real paired nighttime videos remains impractical because rain effects cannot be isolated from other degradations like sensor noise. To bridge this gap, we introduce \textbf{UENR-600K}, a large-scale, physically grounded dataset containing 600,000 1080p frame pairs. We utilize Unreal Engine to simulate rain as 3D particles within virtual environments. This approach guarantees photorealism and physically real raindrops, capturing correct details like color refractions, scene occlusions, rain curtains. Leveraging this high-quality data, we establish a new state-of-the-art baseline by adapting the Wan 2.2 video generation model. Our baseline treat deraining as a video-to-video generation task, exploiting strong generative priors to almost entirely bridge the sim-to-real gap. Extensive benchmarking demonstrates that models trained on our dataset generalize significantly better to real-world videos. 
\end{abstract}

\section{Introduction}

Nighttime video deraining is critical for autonomous surveillance and navigation, yet it remains a major challenge compared to its daytime counterpart \cite{NightRain, RLP}. This difficulty stems from the complex optical properties of nighttime rain that differ significantly from daytime conditions \cite{VisionRain}. First, nighttime rain is \textbf{chromatic}: raindrops refract colored artificial (\eg neon) light rather than appearing as the white streaks typical of daytime rain \cite{VisionRain, GTAVNightRain}. Second, rain visibility is \textbf{localized} near light sources, unlike the uniform patterns in daytime videos \cite{RLP}. Third, nighttime rain exhibits a \textbf{glimmer effect}, producing sudden high-intensity flashes as raindrops pass through focused light \cite{RainRendering}. Fourth, wind-driven rain forms volumetric \textbf{rain curtains} that shift across the scene, visible especially under illumination \cite{OutdoorRain}. These properties are absent in daytime rain, where uniform sunlight instead produces globally visible, colorless streaks.

Current research struggles to address these challenges due to a severe lack of high-quality data \cite{NightRain}. Nighttime models could not be trained using daytime datasets \cite{JORDER, NTURain, LHPRain} because they lack these specific lighting interactions, and single-image datasets \cite{SPAData, RealRain1K, GTAVNightRain} fail to provide either temporal rain dynamics or scene consistency. Furthermore, capturing large-scale real-world paired videos at night is impractical due to prohibitive collection costs and entangled degradations (\eg, severe sensor noise) that make isolating the rain effect impossible \cite{GTRain, NightRain}. Consequently, for nighttime video deraining, existing studies rely on limited synthetic datasets \cite{SynNightRain, ASFNet}. However, these datasets essentially overlay rain as a global layer onto 2D video frames, without modeling scene content or lighting. Therefore, the resulting rain appears as floating, repetitive foggy patterns rather than realistic, localized, chromatic streaks.

To bridge this gap, we present UENR-600K, a large-scale dataset for nighttime video deraining. Unlike previous overlay methods, we utilize Unreal Engine 5 to simulate rain within a virtual raining environment. In this environment, we use cinematic cameras to capture paired videos, in which raindrops and rain curtains are correctly occluded by scene objects and exhibit accurate chromaticity. We generate 600,000 pairs of 1080p frames with extensive diversity. We dynamically control parameters such as wind direction, rain intensity, motion blur, camera parameters, and color grading.

To demonstrate the utility of this dataset, we retrain eight deraining methods on both our dataset and SynNightRain \cite{SynNightRain}, and conduct extensive benchmarking on real and synthetic videos. The models span CNN, RNN, Transformer, and diffusion architectures. We also establish a strong baseline by adapting the Wan 2.2 video generation model for deraining, and evaluate all methods on real nighttime rain videos using a vision-language model (VLM) as judge. Training on our dataset leads to better real-world deraining than SynNightRain for most methods, with our baseline judged best in 94\% of test videos. On real videos, our baseline, by leveraging Wan's strong generative prior, almost entirely bridges the sim-to-real gap, allowing it to effectively handle night rain-specific phenomena such as chromatic rain, localized streaks, and rain-induced fog.

Our contributions are summarized as follows:

\begin{itemize}
    \item \textbf{Dataset.} We propose UENR-600K, the first large-scale, physically grounded dataset for nighttime video deraining, containing 600,000 1080p frame pairs rendered in Unreal Engine.
    \item \textbf{Baseline.} We adapt the Wan 2.2 video generation model into a video-to-video architecture for nighttime video deraining. Finetuned on our dataset, this model achieves a dominating 94\% preference rate on real videos.
    \item \textbf{Benchmarking.} We conduct a comprehensive analysis of eight diverse deraining methods across synthetic and real domains, demonstrating the novelty of our dataset compared to existing synthetic datasets. Furthermore, for most methods, training on our data produces better deraining results, demonstrating that physically grounded rain data improves real-world generalization.
\end{itemize}
\section{Related Works}

% \usepackage{multirow}
% \usepackage{graphicx}
% \usepackage{pifont} % Required for \ding{51}
\begin{table}[t]
\centering
\caption{Overview of deraining datasets. Among the datasets surveyed, our dataset is the only one providing video data of in-scene nighttime rain.}
\label{tab:datasets}
\resizebox{\textwidth}{!}{%
\begin{tabular}{lcccccccccc}
\toprule
\multicolumn{1}{c}{\textbf{Dataset}} & \textbf{Year}         & \textbf{Venue}        & \textbf{\begin{tabular}[c]{@{}c@{}}Rain\\ Type$^a$\end{tabular}} & \textbf{\begin{tabular}[c]{@{}c@{}}GT\\ Type$^b$\end{tabular}} & \textbf{\begin{tabular}[c]{@{}c@{}}\#Rainy\\ Frames$^c$\end{tabular}} & \textbf{\begin{tabular}[c]{@{}c@{}}\#GT\\ Frames$^c$\end{tabular}} & \textbf{Resolution} & \textbf{Night$^d$} & \textbf{Video$^e$} & \textbf{\begin{tabular}[c]{@{}c@{}}Raining\\ In-Scene$^f$\end{tabular}} \\ \midrule
Rain200L/H \cite{JORDER}                           & 2017                  & CVPR                  & Synth.                                                       & Real                                                       & 4,000           & 4,000         & 435$\times$366        &                &                &                                                                      \\
DDN-Data \cite{DDN}                             & 2017                  & CVPR                  & Synth.                                                       & Real                                                       & 13,000          & 13,000        & 489$\times$428        &                &                &                                                                      \\
DID-Data \cite{DID-MDN}                             & 2018                  & CVPR                  & Synth.                                                       & Real                                                       & 13,200          & 13,200        & 512$\times$512        &                &                &                                                                      \\
NTURain$^g$ \cite{NTURain}                              & 2018                  & CVPR                  & Synth.                                                       & Real                                                       & -               & -             & 640$\times$480        &                & \ding{51}      &                                                                      \\
RainSynAll25 \cite{EraseFill}                         & 2018                  & CVPR                  & Synth.                                                       & Real                                                       & -               & -             & -                   &                & \ding{51}      &                                                                      \\
Rain800 \cite{IDCGAN}                              & 2017                  & -                     & Synth.                                                       & Real                                                       & 800             & 800           & 518$\times$419        &                &                &                                                                      \\
SPA-Data \cite{SPAData}                             & 2019                  & CVPR                  & Real                                                         & Derived                                                    & 29,500          & 29,500        & 256$\times$256        & \ding{51}      &                &                                                                      \\
Outdoor-Rain \cite{OutdoorRain}                         & 2019                  & CVPR                  & Synth.                                                       & Real                                                       & 10,500          & 10,500        & 720$\times$480        &                &                &                                                                      \\
RainCityscapes \cite{DAFNet}                       & 2019                  & CVPR                  & Synth.                                                       & Real                                                       & 10,620          & 10,620        & 2048$\times$1024      &                &                &                                                                      \\
Rain13k \cite{MSPFN}                              & 2020                  & CVPR                  & Synth.                                                       & Real                                                       & 13,700          & 13,700        & 482$\times$419        &                &                &                                                                      \\
RainDirection \cite{RainDirectionPaper}                        & 2021                  & ICCV                  & Synth.                                                       & Real                                                       & 3,300           & 3,300         & 1945$\times$1444      &                &                &                                                                      \\
RainDS \cite{RainDS}                               & 2021                  & CVPR                  & Both                                                         & Real                                                       & 5,800           & 5,800         & 818$\times$460        &                &                &                                                                      \\
GT-Rain \cite{GTRain}                              & 2022                  & ECCV                  & Real                                                         & Real                                                       & 31,500          & 31,500        & 666$\times$339        &                &                &                                                                      \\
SynNightRain \cite{SynNightRain}                         & 2022                  & ECCV                  & Synth.                                                       & Real                                                       & 6,000           & 6,000         & 1920$\times$1080      & \ding{51}      & \ding{51}      &                                                                      \\
RealRain-1K \cite{RealRain1K}                          & 2022                  & -                     & Real                                                         & Derived                                                    & 1,120           & 1,120         & 1512$\times$973       & \ding{51}      &                &                                                                      \\
GTAV-NightRain \cite{GTAVNightRain}                       & 2022                  & -                     & Synth.                                                       & Synth.                                                     & 14,146          & 14,146        & 1920$\times$1080      & \ding{51}      &                & \ding{51}                                                            \\
\multirow{2}{*}{LHP-Rain$^h$ \cite{LHPRain}}             & \multirow{2}{*}{2023} & \multirow{2}{*}{ICCV} & \multirow{2}{*}{Real}                                        & \multirow{2}{*}{Derived}                                   & 1,000,000       & 1,000,000     & 256$\times$256        & \ding{51}      &                &                                                                      \\
                                     &                       &                       &                                                              &                                                            & 3,000           & 3,000         & 1920$\times$1080      & \ding{51}      &                &                                                                      \\
HQ-Rain \cite{HQRain}                              & 2023                  & -                     & Synth.                                                       & Real                                                       & 5,000           & 5,000         & 1367$\times$931       &                &                &                                                                      \\
4K-Rain13k \cite{4KRain13k}                           & 2024                  & AAAI                  & Synth.                                                       & Real                                                       & 13,000          & 13,000        & 3840$\times$2160      &                &                &                                                                      \\
RAVD$^i$ \cite{ASFNet}                                & 2025                  & PR                    & Synth.                                                       & Real                                                       & 1,147           & 1,147         & 640$\times$480        & \ding{51}      & \ding{51}      &                                                                      \\
AllWeatherNight \cite{AllWeatherNight}                     & 2026                  & AAAI                  & Both                                                         & Real                                                       & 10,000          & 10,000        & 640$\times$360        & \ding{51}      &                &                                                                      \\ \midrule
UENR-600K (Ours)                        & 2026                  & -                     & Synth.                                                       & Synth.                                                     & 600,000         & 600,000       & 1920$\times$1080      & \ding{51}      & \ding{51}      & \ding{51}                                                            \\ \bottomrule
\multicolumn{11}{l}{\footnotesize $^a$ \textbf{Rain Type}: Synth.\ = synthetically generated, Real = captured in real rain, Both = mixture.} \\
\multicolumn{11}{l}{\footnotesize $^b$ \textbf{GT Type}: the source of clean counterparts; Real = separately captured clean images, Synth.\ = rendered clean} \\
\multicolumn{11}{l}{\footnotesize \quad frames, Derived = estimated from rainy inputs (\textit{e.g.}, via temporal aggregation).} \\
\multicolumn{11}{l}{\footnotesize $^c$ \textbf{\#Rainy/GT Frames}: total number of rainy and clean frames.} \\
\multicolumn{11}{l}{\footnotesize $^d$ \textbf{Night}: includes nighttime rain data. \quad \quad $^e$ \textbf{Video}: provides video sequences (not just individual images).} \\
\multicolumn{11}{l}{\footnotesize $^f$ \textbf{Raining In-Scene}: rain is simulated within a virtual scene where it interacts with scene objects and lighting,} \\
\multicolumn{11}{l}{\footnotesize \quad as opposed to being overlaid onto existing footage.} \\
\multicolumn{11}{l}{\footnotesize $^g$ Contains 4,667 rain videos and 2,677 ground-truth videos. \quad \quad $^h$ Only counted open-sourced partitions.} \\
\multicolumn{11}{l}{\footnotesize $^i$ Only counted the open-sourced partition. This partition does not contain nighttime deraining data.} \\
\end{tabular}%
}
\end{table}

\paragraph{Video deraining} aims to remove rain streaks and raindrops from video sequences while preserving scene content and temporal consistency. Existing methods address this task through supervised temporal modeling \cite{NTURain, EraseFill, ESTINet, RDDNet}, self-supervised or semi-supervised learning \cite{SelfLearningDerain, DynRainGen, S2VD}, and general video restoration architectures \cite{Turtle}. Notably, nearly every method introduces a new dataset tailored to its target scenario \cite{NTURain, EraseFill, RDDNet, SynNightRain, NightRain, ASFNet}, reflecting the tight coupling between data quality and model performance. This pattern highlights the critical role of training data, particularly for underexplored conditions such as nighttime.

\paragraph{Nighttime deraining datasets.} While numerous datasets (summarized in Tab.~\ref{tab:datasets}) target daytime deraining \cite{DDN, JORDER, DID-MDN, SPAData, OutdoorRain, MSPFN, GTRain, HQRain, 4KRain13k, RealRain1K}, nighttime data remain scarce \cite{NightRain}. Many nighttime data contain images only \cite{SPAData, RealRain1K, GTAVNightRain}, which cannot be used to train video models. Existing nighttime video datasets include SynNightRain \cite{SynNightRain}, RealRain-1K \cite{RealRain1K}, and RAVD \cite{ASFNet}, with only SynNightRain open-sourced nighttime video data. Although RAVD \cite{ASFNet} uses a Direct3D particle system for rain rendering, essentially, all three datasets are constructed by overlaying synthesized rain globally onto the original footage. Such overlayed rain could not respond to scene objects or artificial lighting. As a result, they fail to produce nighttime rain properties such as chromaticity (raindrops refracting colored artificial light) or the glimmer effect (sudden high-intensity flashes near focused light sources) \cite{NightRain, GTAVNightRain, RLP}. These properties are also why daytime datasets cannot be used for nighttime settings. To align with nighttime rain properties, GTAV-NightRain~\cite{GTAVNightRain} explore using a game engine (GTA-V) to capture paired data by toggling rain, but this dataset contains only 14,146 image pairs with no video data. 

Furthermore, capturing real-world paired nighttime video is fundamentally impractical for two reasons: first, collecting large-scale paired videos is prohibitively costly; second, nighttime environments introduce entangled degradations (such as severe sensor noise), making it nearly impossible to isolate the rain effect. To date, no large-scale, physically grounded nighttime video deraining dataset exists, constraining progress in this field. 

\paragraph{Video generation and editing.} Beyond paired training data, generative priors from large-scale models offer another avenue for restoration. Diffusion-based video generation models \cite{Sora, Wan, SVD, CogVideoX} synthesize videos from text or image inputs and learn strong priors about scene appearance and motion dynamics. These models, however, are designed for generation rather than editing; adapting them for specific restoration tasks such as deraining \cite{WeatherDiff, DiffBIR, UpscaleAVideo} requires high-quality, physically grounded paired data to steer their capabilities correctly. This further motivates our dataset construction.

\section{The UENR-600K Dataset}
\label{sec:dataset}

\subsection{Defining Nighttime Video Deraining}
\label{sec:dataset_definition}

\begin{figure}[t]
    \centering
    \includegraphics[width=\linewidth]{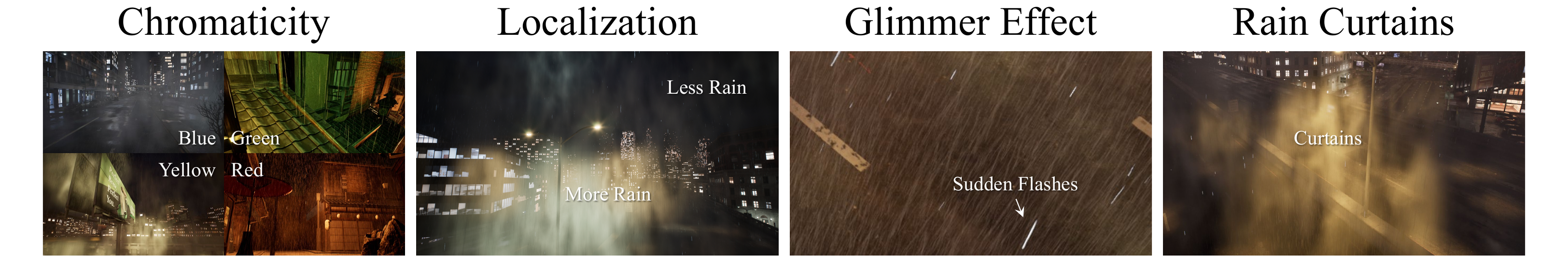}
    \caption{Properties of nighttime rain illustrated with frames from our dataset. \textbf{Chromaticity:} raindrops refract colored artificial light (blue, yellow, green, red) rather than appearing white. \textbf{Localization:} rain is visible near light sources but fades in unlit regions. \textbf{Glimmer effect:} raindrops produce sudden high-intensity flashes as they pass through focused light beams. \textbf{Rain curtains:} wind-driven sheets of rain form volumetric, shifting patterns.}
    \label{fig:night_rain_properties}
\end{figure}

Nighttime rain is fundamentally more complex than daytime rain due to interactions with scene geometry and artificial lighting. We define the unique features of nighttime rain across four key aspects. \textbf{Chromaticity:} Raindrops act as refractive lenses. For example, rain passing near colored light sources like neon signs correctly refracts the local color rather than appearing white. \textbf{Localization:} Rain visibility is highly dependent on artificial light sources, such as the light cones of street lamps. \textbf{Glimmer Effect:} As raindrops rapidly pass through focused beams of light, they exhibit sudden and high-intensity flashes. \textbf{Rain Curtains:} Wind and heavy precipitation create shifting and volumetric sheets of rain. Based on these features, we define \textit{nighttime video deraining} as the removal of rain streaks and rain-induced effects, including rain curtains, rain-induced fogs, and the glimmer effect, while ensuring all other video content remains unchanged.

\subsection{Rain Simulation and Rendering}
\label{sec:dataset_simulation}

To capture these nighttime rain phenomena, we abandon the standard practice of 2D rain overlays \cite{SynNightRain}. Instead, we utilize Unreal Engine 5 to construct a true 3D virtual raining environment. In this environment, rain exists as physical particles rather than a global layer. This particle-based approach guarantees physical realism. Because the rain exists in the 3D space, raindrops are naturally occluded by scene objects. Furthermore, the rain dynamically responds to the environment by refracting nearby artificial light colors and becoming illuminated by specific light sources to produce localized rain effects. We capture the paired clear and raining videos within this environment using virtual cinematic cameras.

\subsection{Dataset Diversity and Parameterization}
\label{sec:dataset_diversity}

To prevent models from overfitting to specific camera views or specific types of rain, we introduce extensive variance into our rendering pipeline. We generate continuous camera movement through the 3D environments to capture numerous perspectives, including tracking, aerial, and backward-facing angles. This camera motion also introduces temporal dynamics to the captured video. Alongside camera movement, we randomize variables across three main categories:
\begin{itemize}
    \item \textbf{Weather dynamics:} Rain intensity, rain curtain density, gust amount, wind speed, and wind direction.
    \item \textbf{Rain streak optical effects:} Motion blur intensity and glimmer intensity.
    \item \textbf{Cinematography:} Aperture (\textit{f}/2.8 to \textit{f}/5.6), focal length (8mm to 120mm), and white balance (3500K to 6000K).
\end{itemize}

\begin{figure}[t]
    \centering
    \includegraphics[width=\linewidth]{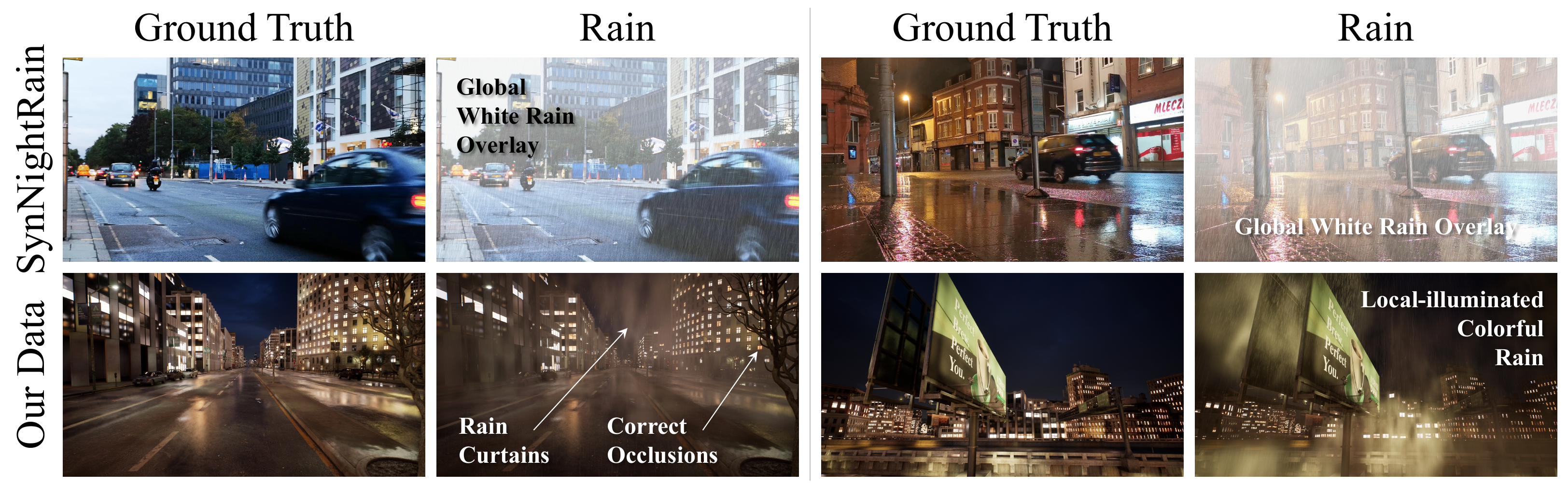}
    \caption{Comparison of rain synthesis between SynNightRain \cite{SynNightRain} (top row) and our dataset (bottom row). Each pair shows a ground-truth frame alongside its rainy counterpart. SynNightRain overlays rain as a global white layer that uniformly covers the entire frame, without responding to scene geometry or lighting. Our dataset simulates rain within a virtual scene: raindrops are correctly occluded by scene objects, form volumetric rain curtains, and refract local artificial light to produce colorful, spatially varying streaks.}
    \label{fig:dataset_comparison}
\end{figure}

Using this pipeline, we render a total of 600,000 paired frames divided into two subsets. The primary subset "City Sample" contains 500,000 frames captured as a single continuous camera trajectory to provide extensive view diversity. The secondary subset "Kyoto" contains 100,000 frames captured in a detailed alleyway environment with colorful illuminations to provide additional variations.

\section{Methodology}

\subsection{Model Architecture}
\label{sec:methodology_architecture}

\begin{figure}[t]
    \centering
    \includegraphics[width=\linewidth]{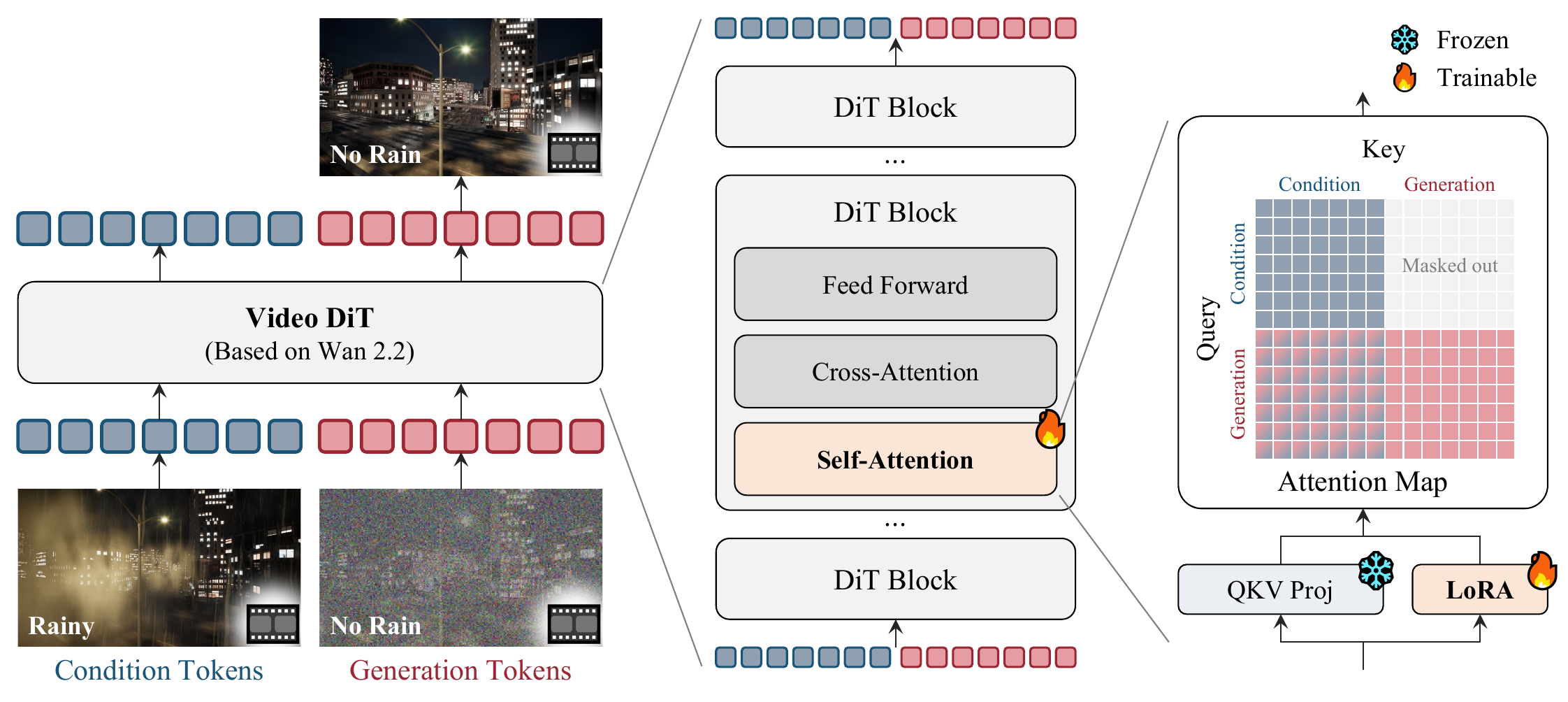}
    \caption{Our baseline architecture, adapted from the Wan 2.2 Video DiT. The rainy input is encoded into condition tokens (blue) and concatenated with generation tokens (red); the DiT denoises only the generation tokens while using the condition tokens as context. A unidirectional attention mask prevents condition tokens from attending to generation tokens, keeping the input uncorrupted. Only LoRA adapters on the QKV projections are trained; all other parameters stay frozen.}
    \label{fig:model_architecture}
\end{figure}

We establish a baseline for nighttime video deraining by adapting the Wan 2.2 video generation model \cite{Wan} into a video-to-video architecture. As illustrated in Fig. \ref{fig:model_architecture}, we begin by encoding the input rainy video into a sequence of \textit{condition tokens}. We then concatenate these condition tokens with a parallel sequence of \textit{generation tokens}. To ensure strict spatial and temporal correspondence between the input and output videos, we apply identical positional embeddings to both sets of tokens. During the subsequent denoising process, the model must not corrupt the input condition. To ensure this, we introduce a unidirectional mask into the self attention maps of the diffusion transformer blocks. This mask actively prevents the condition tokens from attending to the generation tokens. Finally, we isolate the denoised generation tokens and decode them to produce the clean video output.

\subsection{Finetuning and Inference Objectives}
\label{sec:methodology_finetuning}

We finetune this model using a flow matching objective \cite{Wan}. This objective trains a network $v_\theta$ to predict a ground truth velocity $v_t$. We define this velocity along a linear path $x_t$ interpolating between pure noise $x_0 \sim \mathcal{N}(0, I)$ and the clean latent tokens $x_1$ \cite{rectifiedflow}. Because the path $x_t = tx_1 + (1-t)x_0$ is linear, the ground truth $v_t$ is the exact derivative of the path:

\begin{equation}
    v_t = \frac{dx_t}{dt} = x_1 - x_0.
\end{equation}

\noindent We optimize the model using a mean squared error loss to predict this exact velocity. Let $x^\text{con}$ denote condition tokens (input video) and $x^\text{gen}$ denote generation tokens (output video), and $c_\text{text}$ be a fixed text prompt's embedding ("No rain video"). The loss function is then defined as:

\begin{equation}
    L = \mathbb{E}_{x_1^\text{con}, x_t^\text{gen}, t} \lVert v_\theta(x_1^\text{con}, x_t^\text{gen}, c_\text{text}, t) - v_t  \rVert^2.
\end{equation}

\noindent During inference, we produce the final derained video by integrating the predicted velocity from $t=0$ to $1$:

\begin{equation}
    x_1^\text{gen} = x_0 + \int_0^1 v_\theta(x_1^\text{con}, x_t^\text{gen}, c_\text{text}, t) dt.
\end{equation}

% \begin{figure*}
%     \centering
%     \includegraphics[width=0.92\linewidth]{figures/methodology_model.pdf}
%     \caption{The proposed baseline architecture. We adapt Wan 2.2 for video deraining by encoding the rainy input video into condition tokens (blue). We concatenate these with generation tokens (red). A unidirectional self attention mask prevents condition tokens from attending to generation tokens. Finally, we decode only the generation tokens to output the clean video.}
%     \label{fig:methodology_model}
% \end{figure*}
\section{Experiment}

\subsection{Experiment Setup}

\subsubsection{Datasets and Deraining Models}

\paragraph{Datasets.} We train all methods on two nighttime rain datasets: our UENR-600K (the City Sample subset, 498,200 training / 1,800 test frames) and SynNightRain \cite{SynNightRain} (11,915 / 1,600 frames), the only publicly available nighttime video deraining dataset (Tab.~\ref{tab:datasets}). For real-world evaluation without ground truth, following \cite{NightRain}, we collect 124 nighttime rain videos (11,160 frames) from Pexels \cite{Pexels}, a large-scale video sharing platform whose diverse contributor base provides footage videos, including diverse real nighttime rain scenes. All frames are at 1280$\times$720 resolution. Dataset splits are summarized in Tab.~\ref{tab:dataset_splits} (Appendix~\ref{sec:appendix_data}).

\paragraph{Baseline deraining models.} We train eight methods on both datasets, spanning four architecture families. \textbf{CNN-based:} UConNet \cite{UConNet} and RDD-Net \cite{RDDNet}. \textbf{RNN-based:} ESTINet \cite{ESTINet}. \textbf{Transformer-based:} RLP \cite{RLP} and Turtle \cite{Turtle}. \textbf{Diffusion-based:} WeatherDiff \cite{WeatherDiff}, NightRain \cite{NightRain}, and our baseline. The seven existing methods learn deraining from scratch, while our baseline finetunes a pretrained video generation model. All methods take video inputs except WeatherDiff, which processes single images.

\paragraph{Training and inference.} All methods follow their original training configurations, with total gradient updates matched for fair comparison. Existing methods process fixed-size patches via non-overlapping sliding windows; our baseline operates at full 1280$\times$720 resolution on 90-frame clips without patch decomposition. Full details are in Appendix~\ref{sec:appendix_training}.

\subsubsection{Evaluation and Metrics}

\paragraph{Evaluating real nighttime rain deraining.} We collect 124 real nighttime rain videos from Pexels \cite{Pexels} (90 frames each, 1280$\times$720) and run all eight methods on them. Since no ground truth exists, we use a vision-language model (VLM; \texttt{claude-sonnet-4-6} \cite{Claude}) as an automated judge. Given output frames from competing methods, the VLM selects the best result based on rain removal, detail preservation, artifact absence, and overall quality, with a required written justification. Method labels are randomized per evaluation to avoid position bias (prompt used and full protocol are in Appendix~\ref{sec:appendix_vlm}).

\paragraph{Evaluating synthetic nighttime rain deraining.} On synthetic test sets (1,800 frames for our dataset, 1,600 for SynNightRain), ground-truth clean frames are available. We report PSNR (pixel-level fidelity, in dB) and SSIM (structural similarity) in RGB space between derained outputs and ground-truth frames.

\paragraph{Evaluating video temporal consistency.} We measure temporal consistency with two complementary metrics. \textbf{Average frame difference (AFD)} computes the mean LPIPS \cite{LPIPS} between all pairs of consecutive output frames; lower values indicate smoother transitions. \textbf{VLM temporal rating} uses a VLM (\texttt{claude-opus-4-6} \cite{Claude}) to rate each method's output on a 1 to 5 scale (5 = best). For each video clip, the VLM views four consecutive output frames together with pre-computed inter-frame difference maps that highlight pixel changes between adjacent frames. Bright regions in the difference maps reveal flickering or inconsistent rain removal, while a temporally consistent output produces mostly dark difference maps (full protocol in Appendix~\ref{sec:appendix_temporal}). All VLM evaluations costed \$225.55 in total (Tab.~\ref{tab:vlm_cost}).

\subsection{Dataset Comparison}
\label{sec:dataset_comparison}

\paragraph{Training on our dataset improves real-world deraining.} For each method, we train two versions (one on our dataset, one on SynNightRain) and ask a VLM \cite{Claude} to select the better result on real nighttime rain videos (Tab.~\ref{tab:vlm_eval_per_method}). Training on our dataset yields preferred results for all CNN, RNN, and Transformer methods, as well as our diffusion-based baseline. The two exceptions, WeatherDiff and NightRain, are diffusion models operating on 64$\times$64 inputs that cover less than 1\% of a 1280$\times$720 frame (see Input Size column). Without sufficient spatial context, the two models cannot distinguish localized rain phenomena (such as rain curtains near light sources) from clean dark regions, producing spurious white haze at inference; SynNightRain's globally uniform rain does not create this ambiguity. For all methods with sufficient receptive fields, training on our dataset consistently improves real-world deraining across all architecture families. Additionally, models trained on SynNightRain tend to suppress scene content in dark regions, because SynNightRain overlays rain as a global bright layer and models learn to remove brightness across the entire frame.

\begin{figure}
    \centering
    \includegraphics[width=\linewidth]{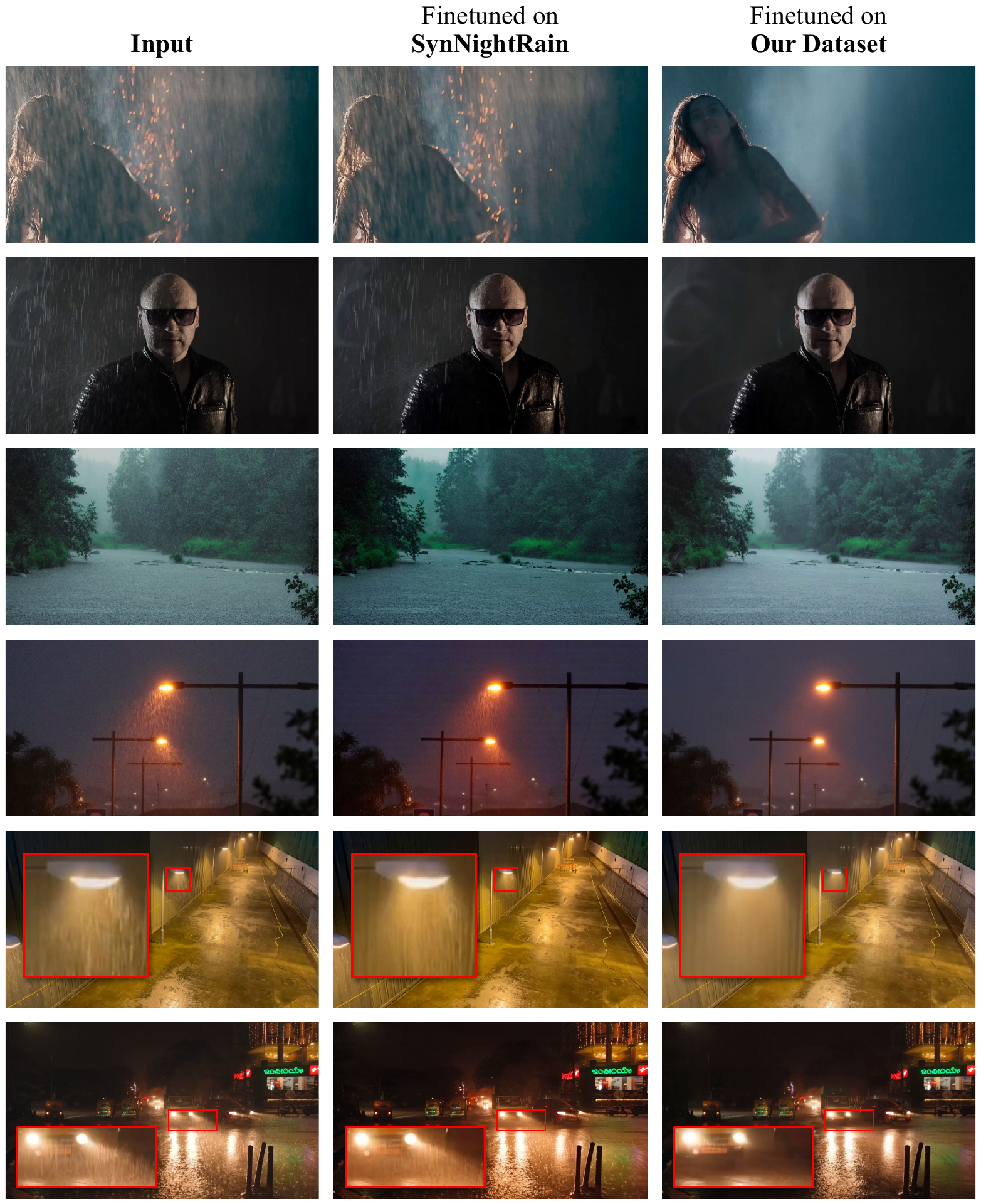}
    \caption{Our baseline finetuned on SynNightRain (middle) versus our dataset (right) across six real nighttime rain scenes from Pexels \cite{Pexels}. The SynNightRain-finetuned baseline struggles with chromatic rain near colored lighting (row 1), localized rain around streetlamps (row 4), and rain-induced haze (rows 5 and 6; see red crops). The version finetuned on our dataset handles all these conditions effectively.}
    \label{fig:wan_on_both_datasets}
\end{figure}

\paragraph{Our dataset enables the baseline to handle diverse nighttime rain conditions.} Fig.~\ref{fig:wan_on_both_datasets} compares our baseline finetuned on each dataset across six real scenes. The baseline finetuned on SynNightRain struggles with chromatic rain near colored lighting (row 1), localized rain around streetlamps (row 4), and rain-induced fog under direct illumination (rows 5 and 6; see red crops). The version finetuned on our dataset handles all these conditions effectively. Our dataset simulates these nighttime-specific phenomena through 3D particle rendering (Sec.~\ref{sec:dataset_simulation}), while SynNightRain does not model nighttime rain properties like chromaticity or localization.

% \usepackage{graphicx}
\begin{table}[t]
\centering
\caption{Pairwise comparison of deraining results on real nighttime rain videos, where a VLM selects the better result based on rain removal and detail preservation. For each method, the version trained on our dataset competes against the version trained on SynNightRain \cite{SynNightRain}; bold indicates the preferred training dataset. \textbf{Input Size} is the spatial resolution each method processes during training and inference. Training on our dataset yields preferred results for all methods with input sizes $\geq$128$\times$128, covering all architecture families. The two exceptions, WeatherDiff and NightRain, use 64$\times$64 inputs (less than 1\% of a 1280$\times$720 frame), which limits their ability to distinguish localized rain from clean dark regions in our data.}
\label{tab:vlm_eval_per_method}
\fontsize{8pt}{10pt}\selectfont
% \resizebox{\textwidth}{!}{%
\begin{tabular}{lc|cc >{\columncolor{gray!20}}c}
\toprule
                      & \textbf{Input Size} & \textbf{SynNightRain \cite{SynNightRain}} & \textbf{Tie} & \textbf{Our Data}   \\ \midrule
\textbf{ESTINet \cite{ESTINet}}  & 224$\times$224 & 16.4\%                & 6.5\%        & \textbf{77.2\%} \\
\textbf{RDD-Net \cite{RDDNet}}      & 128$\times$128 & 32.3\%                & 1.6\%        & \textbf{66.1\%} \\
\textbf{RLP \cite{RLP}}          & 256$\times$256 & 27.7\%                & 8.6\%        & \textbf{63.7\%} \\
\textbf{Turtle \cite{Turtle}}       & 128$\times$128 & 16.1\%                & 6.2\%        & \textbf{77.7\%} \\
\textbf{UConNet \cite{UConNet}}      & 128$\times$128 & 3.2\%                 & 1.3\%        & \textbf{95.4\%} \\
\textbf{WeatherDiff \cite{WeatherDiff}}  & 64$\times$64 & \textbf{91.7\%}       & 3.0\%        & 5.4\%           \\
\textbf{NightRain \cite{NightRain}}    & 64$\times$64 & \textbf{85.2\%}       & 1.6\%        & 13.2\%          \\
\textbf{Our Baseline} & Full & 19.6\%                & 15.1\%       & \textbf{65.3\%} \\ \midrule
\textbf{Average}      & -- & 36.5\%                & 5.5\%        & \textbf{58.0\%} \\ \bottomrule
\end{tabular}%
% }
\end{table}

% \usepackage{graphicx}
\begin{table}[t]
\centering
\caption{Multi-way comparison of deraining results on real nighttime rain videos, where a VLM selects the best result among all eight methods trained on the same dataset. \textbf{Pretrained}: model was pretrained on large-scale video data before finetuning. Our baseline, the only pretrained method, dominates in both settings: 94.4\% when trained on our dataset and 69.4\% when trained on SynNightRain.}
\label{tab:vlm_eval_per_dataset}
\fontsize{8pt}{10pt}\selectfont
% \resizebox{\textwidth}{!}{%
\begin{tabular}{lcc|cc|c}
\toprule
\textbf{}            & \textbf{Architecture} & \textbf{Pretrained} & \textbf{SynNightRain \cite{SynNightRain}} & \textbf{Our Data}   & \textbf{Average} \\ \midrule
\textbf{ESTINet \cite{ESTINet}} & CNN+RNN &  & 12.9\%                & 5.6\%           & 9.3\%            \\
\textbf{RDD-Net \cite{RDDNet}}     & CNN &  & 10.5\%                & 0.0\%           & 5.2\%            \\
\textbf{RLP \cite{RLP}}         & Transformer &  & 4.0\%                 & 0.0\%           & 2.0\%            \\
\textbf{Turtle \cite{Turtle}}      & Transformer &  & 0.8\%                 & 0.0\%           & 0.4\%            \\
\textbf{UConNet \cite{UConNet}}     & CNN &  & 0.0\%                 & 0.0\%           & 0.0\%            \\
\textbf{WeatherDiff \cite{WeatherDiff}} & Diffusion &  & 1.6\%                 & 0.0\%           & 0.8\%            \\
\textbf{NightRain \cite{NightRain}}   & Diffusion &  & 0.8\%                 & 0.0\%           & 0.4\%            \\
\rowcolor{gray!20}
\textbf{Our Baseline}        & Diffusion & \ding{51} & \textbf{69.4\%}       & \textbf{94.4\%} & \textbf{81.9\%}  \\ \bottomrule
\end{tabular}%
% }
\end{table}

\begin{figure}
    \centering
    \includegraphics[width=\linewidth]{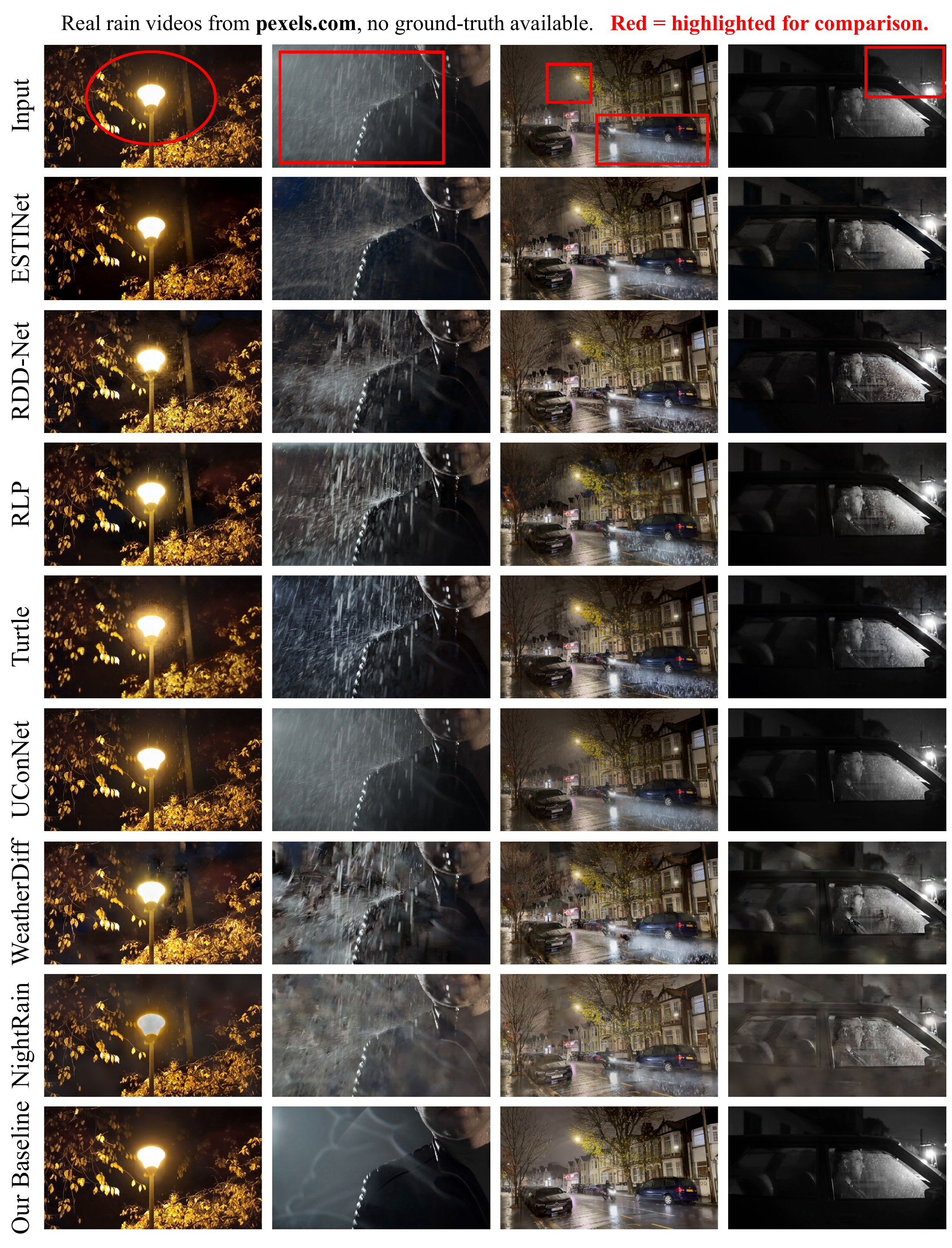}
    \caption{Qualitative comparison of all eight methods on four real nighttime rain scenes. All methods trained on our dataset. Red annotations highlight regions for comparison. Existing restoration methods (ESTINet through UConNet) reduce rain to varying degrees but leave visible streaks in heavy rain regions. The two 64$\times$64 diffusion models (WeatherDiff, NightRain) introduce haze or darken the scene. Our baseline, leveraging its pretrained generative prior, removes rain almost completely without introducing artifacts, demonstrating that the sim-to-real gap is effectively bridged.}
    \label{fig:method_comparison_main}
\end{figure}

\subsection{Deraining Method Comparison}
\label{sec:method_comparison}

\begin{table}[t]
\centering
\caption{Temporal consistency of deraining results on real nighttime rain videos, evaluated by two complementary metrics. \textbf{AFD}$\downarrow$: average inter-frame LPIPS ($\times 10^{2}$; lower is better), measuring perceptual difference between consecutive output frames. \textbf{VLM}$\uparrow$: VLM-rated temporal smoothness on a 1 to 5 scale (higher is better). Our baseline achieves the best temporal consistency under both metrics and both training datasets.}
\label{tab:temporal_consistency}
\fontsize{8pt}{10pt}\selectfont
% \resizebox{\textwidth}{!}{%
\begin{tabular}{l @{\hskip 8pt} cc @{\hskip 10pt} cc @{\hskip 10pt} cc}
\toprule
& \multicolumn{2}{c}{\textbf{SynNightRain}} & \multicolumn{2}{c}{\textbf{Our Data}} & \multicolumn{2}{c}{\textbf{Average}} \\
\cmidrule(lr){2-3} \cmidrule(lr){4-5} \cmidrule(lr){6-7}
& AFD$\downarrow$ & VLM$\uparrow$ & AFD$\downarrow$ & VLM$\uparrow$ & AFD$\downarrow$ & VLM$\uparrow$ \\ \midrule
\textbf{ESTINet \cite{ESTINet}}        & \textbf{6.40} & 3.22 & 5.70 & 3.32 & 6.05 & 3.27 \\
\textbf{RDD-Net \cite{RDDNet}}         & 7.04 & 3.37 & 9.93 & 2.37 & 8.49 & 2.87 \\
\textbf{RLP \cite{RLP}}                & 7.75 & 3.11 & 10.11 & 2.60 & 8.93 & 2.86 \\
\textbf{Turtle \cite{Turtle}}          & 8.55 & 2.71 & 8.08 & 3.07 & 8.32 & 2.89 \\
\textbf{UConNet \cite{UConNet}}        & 7.94 & 2.47 & 9.34 & 3.03 & 8.64 & 2.75 \\
\textbf{WeatherDiff \cite{WeatherDiff}} & 10.56 & 2.13 & 22.79 & 1.04 & 16.68 & 1.59 \\
\textbf{NightRain \cite{NightRain}}    & 7.75 & 2.63 & 9.13 & 2.03 & 8.44 & 2.33 \\
\rowcolor{gray!20}
\textbf{Our Baseline}                  & 6.88 & \textbf{3.56} & \textbf{2.47} & \textbf{4.46} & \textbf{4.68} & \textbf{4.01} \\ \bottomrule
\end{tabular}%
% }
\end{table}

\paragraph{Quantitatively, our baseline achieves the best deraining preference among all methods.} We conduct a VLM-driven multi-way comparison where all eight methods, trained on the same dataset, compete on each real nighttime rain video. The VLM selects the best result (Tab.~\ref{tab:vlm_eval_per_dataset}). Our baseline dominates both settings: 94.4\% preference when trained on our dataset and 69.4\% on SynNightRain. The performance is followed by ESTINet and RDD-Net (9.3\% and 5.2\% average preference), and then the remaining five methods (below 2\%). Our baseline, as the only pretrained method, benefits from both its strong generative prior and our physically grounded data, which together raise the preference rate from 69.4\% to 94.4\%.

\paragraph{Qualitatively, on real videos, most methods still leave visible rain or introduce artifacts.} Fig.~\ref{fig:method_comparison_main} compares all eight methods trained on our dataset. Restoration methods (ESTINet, RDD-Net, RLP, Turtle, UConNet) reduce rain but leave visible streaks in heavy-rain regions such as under streetlamps (column 1) and in close-up rain (column 2), because they learn pixel-level mappings without a generative prior to hallucinate missing content. The two 64$\times$64 diffusion models behave differently: WeatherDiff introduces severe patchy artifacts from independently denoising small tiles, while NightRain darkens the scene and loses detail in already dim regions (column 4). Our baseline removes rain almost completely across all four scenes, including chromatic streaks and rain-induced fog, without suppressing scene detail or introducing artifacts.

\paragraph{Our baseline also achieves the best temporal consistency.} Tab.~\ref{tab:temporal_consistency} reports AFD and VLM temporal ratings for all methods. Our baseline achieves the lowest AFD and highest VLM rating (AFD 2.47, VLM 4.46 when trained on our data). The key distinction is whether a method processes multiple frames jointly or each frame independently. Single-image methods and small-patch diffusion models score worst because each frame or patch is processed independently, producing visible flickering. Our baseline achieves both the best deraining quality and the smoothest temporal output.

\subsection{Analysis on Synthetic Data}
\label{sec:quantitative_synthetic}

% \usepackage{multirow}
% \usepackage{graphicx}
\begin{table}[t]
\centering
\caption{Same-dataset benchmarking: each method is trained and tested on the same dataset. \ensuremath{\Delta}PSNR = Our Data PSNR minus SynNightRain PSNR; positive values indicate that SynNightRain is easier to fit, while the negative value (bold) indicates the opposite. All seven existing methods achieve higher PSNR on SynNightRain, confirming its simpler degradation patterns. Our baseline is the only method that scores higher on our dataset, because its pretrained generative prior aligns better with rendering-based rain.} % 
\label{tab:same_dataset}
\fontsize{8pt}{10pt}\selectfont
% \resizebox{\textwidth}{!}{%
\begin{tabular}{lccccccc}
\toprule
\multicolumn{1}{c}{\textbf{}} & \multicolumn{2}{c}{\textbf{Our Data}} & \textbf{} & \multicolumn{2}{c}{\textbf{SynNightRain \cite{SynNightRain}}} & \textbf{} & \multirow{2}{*}{\textbf{\ensuremath{\Delta}PSNR}} \\ \cmidrule(lr){2-3} \cmidrule(lr){5-6}
\multicolumn{1}{c}{\textbf{}} & \textbf{PSNR$\uparrow$}     & \textbf{SSIM$\uparrow$}     & \textbf{} & \textbf{PSNR$\uparrow$}       & \textbf{SSIM$\uparrow$}       & \textbf{} &                                      \\ \midrule
\textbf{ESTINet \cite{ESTINet}}              & 24.88             & 0.845             &           & 26.89               & 0.867               &           & $-$2.01                                 \\
\textbf{RDD-Net \cite{RDDNet}}              & 24.34             & 0.835             &           & 25.03               & 0.851               &           & $-$0.69                                 \\
\textbf{RLP \cite{RLP}}                  & 24.34             & 0.833             &           & 25.23               & 0.835               &           & $-$0.89                                 \\
\textbf{Turtle \cite{Turtle}}               & 19.35             & 0.712             &           & 24.40                & 0.838               &           & $-$5.05                                 \\
\textbf{UConNet \cite{UConNet}}              & 18.82             & 0.713             &           & 21.85               & 0.778               &           & $-$3.03                                 \\
\textbf{WeatherDiff \cite{WeatherDiff}}          & 23.88             & 0.848             &           & 27.52               & 0.896               &           & $-$3.64                                 \\
\textbf{NightRain \cite{NightRain}}            & 16.50              & 0.667             &           & 19.03               & 0.777               &           & $-$2.53                                 \\
\textbf{Our Baseline}         & 25.49             & 0.834             &           & 23.31               & 0.755               &           & $+$2.18                                 \\ \bottomrule
\end{tabular}%
% }
\end{table}

\paragraph{Our dataset is more challenging for existing methods, but easier for finetuning our baseline.} As shown in Tab.~\ref{tab:same_dataset}, existing methods achieve an average of 2.55 dB higher PSNR on SynNightRain than on our dataset, indicating SynNightRain's algorithmic rain presents a simpler degradation pattern to fit. In contrast, our baseline achieves a 2.18 dB higher PSNR on our dataset. This divergence occurs because our baseline builds upon a video generation model pretrained on large-scale natural video \cite{Wan}. With this strong generative prior, our baseline inherently understands real physical interactions and lighting, and adapts easily to our physically simulated rain. Conversely, SynNightRain's algorithm-generated rain overlay deviates from natural physics, making it harder for the generative model to adapt. This performance difference validates the superior realism of our proposed data.

\begin{figure}[t]
    \centering
    \includegraphics[width=\linewidth]{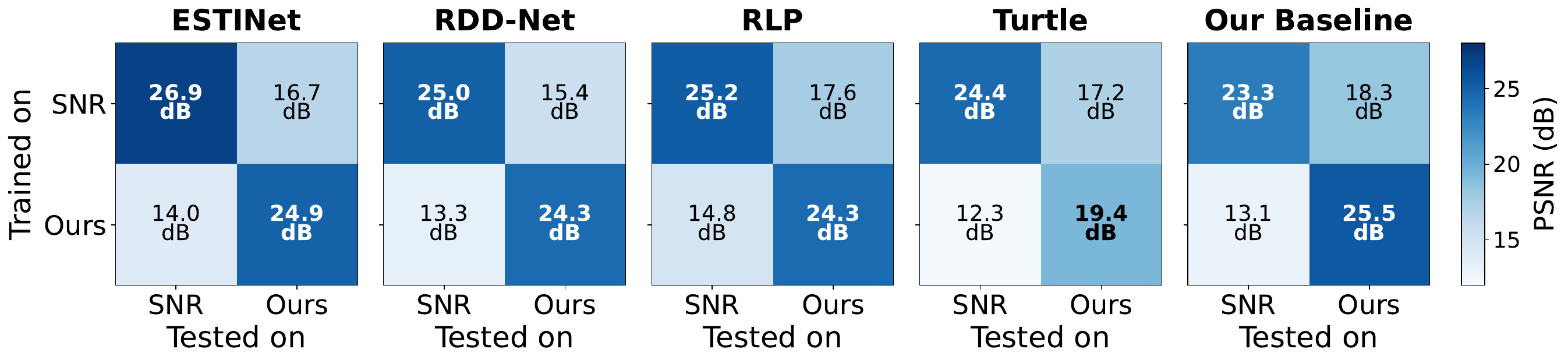}
    \caption{Cross-dataset evaluation of five representative methods. Each method is trained on one dataset and tested on both. Diagonal cells represent same-dataset performance, while off-diagonal cells show cross-dataset performance in PSNR. All methods experience a substantial performance drop when tested on a different dataset than their training source, highlighting the significant domain gap between our physically grounded data and the algorithmically synthesized SynNightRain.}
    \label{fig:cross_dataset_confusion}
\end{figure}

\paragraph{Cross-dataset evaluation shows a large domain gap, highlighting our data's novelty.} To measure the domain gap between datasets, we train the models on one dataset and evaluate them on both. In Fig.~\ref{fig:cross_dataset_confusion}, the off-diagonal cells display significantly lower PSNR values (7 to 11 dB drop) when testing across datasets. This mutual failure confirms a large domain gap: models optimized for the 2D overlays of SynNightRain cannot handle our physically grounded 3D rain, and vice versa. This contrast establishes the novelty of UENR-600K as a fundamentally different degradation domain that existing synthetic nighttime rain data does not capture.
% \paragraph{Model rankings are consistent across evaluations.} The synthetic data results (PSNR/SSIM) from Tab.~\ref{tab:same_dataset} show the same tiers as the VLM preference results: our baseline ranks first in both, followed by ESTINet and RDD-Net as the strongest existing methods, and then the rest. This agreement between pixel-level metrics and VLM-based visual preference judgments indicates that our benchmark produces stable and reliable rankings.
\section{Conclusion}

We presented UENR-600K, a large-scale physically real nighttime deraining dataset rendered in Unreal Engine. Training on our data generalizes much better to real-world nighttime deraining, with our introduced new baseline achieving the best performance. 

\paragraph{Limitations and Future Analysis.} \textbf{(1)} We did not tune hyperparameters for our baseline; systematic tuning may further improve performance. \textbf{(2)} The Kyoto subset (100,000 frames) remains unused. It offers additional scene diversity with dense colorful illuminations, which could benefit future work's training and benchmarking.

\bibliographystyle{splncs04}
\bibliography{main}

\clearpage

\clearpage
\setcounter{page}{1}
\title{Supplementary Material for\\UENR-600K: A Large-Scale Physically Grounded Dataset for Nighttime Video Deraining} 
\titlerunning{Supplementary Material for UENR-600K}
\author{}
\authorrunning{}
\institute{}
\maketitle

\vspace{-20pt}

% \begin{figure}
%     \centering
%     \includegraphics[width=\linewidth]{figures/project_page_eyecatching.pdf}
% \end{figure}

\setcounter{section}{0}
\renewcommand{\thesection}{\Alph{section}}

\section{Data Processing Details}
\label{sec:appendix_data}

\paragraph{Our dataset.} We use the City Sample subset of UENR-600K, which consists of 500,000 frames rendered as a single continuous camera trajectory in Unreal Engine 5. All frames are pre-rendered at 1280$\times$720 resolution. We designate frames 0 through 1,799 (1,800 frames) as the test set and frames 1,800 through 500,000 (498,200 frames) as the training set. Because the frames come from a continuous trajectory, adjacent frames maintain temporal coherence and can be grouped for video models. For testing, we segment the 1,800 test frames into 20 non-overlapping clips of 90 frames each.

\begin{table}[t]
\centering
\caption{Dataset splits used in our experiments. All frames are at 1280$\times$720. UENR-600K and SynNightRain provide paired rainy/clean frames for training and synthetic evaluation. The Pexels set has no ground truth and is used exclusively for VLM-based evaluation.}
\label{tab:dataset_splits}
\fontsize{8pt}{10pt}\selectfont
\begin{tabular}{lccc}
\toprule
& \textbf{UENR-600K (Ours)} & \textbf{SynNightRain \cite{SynNightRain}} & \textbf{Pexels \cite{Pexels}} \\ \midrule
Rain type & Synth. & Synth. & Real \\
Total frames & 600,000$^\dagger$ & 13,515 & 11,160 \\
Train frames & 498,200 & 11,915 & -- \\
Test frames & 1,800 (20 clips) & 1,600 (8 clips) & 11,160 (124 clips) \\
Ground truth & \ding{51} & \ding{51} & -- \\
\bottomrule
\multicolumn{4}{l}{\footnotesize $^\dagger$ 600,000 frames across two subsets (City Sample: 500,000; Kyoto: 100,000).} \\
\multicolumn{4}{l}{\footnotesize \quad Only the City Sample subset is used in this benchmark.} \\
\end{tabular}
\end{table}

\paragraph{SynNightRain.} SynNightRain \cite{SynNightRain} contains 68 videos across two subsets: NightRain (34 videos) and NightRainVeiling (34 videos). Each video contains 200 frames with paired rainy inputs and ground-truth clean frames. The videos have mixed native resolutions (3840$\times$2160, 1920$\times$1080, 1280$\times$720, and 1080$\times$1920 portrait). We follow the original train/test split: videos 002, 003, 005, and 007 from both subsets (8 videos, 1,600 frames) form the test set, and the remaining 60 videos (11,915 frames) form the training set. All frames are resized to 1280$\times$720 using bicubic interpolation.

\paragraph{Real-world evaluation set.} We collect 124 real-world nighttime rain videos from Pexels \cite{Pexels} for out-of-domain evaluation. Each video is resized to 1280$\times$720 and trimmed to 90 frames, yielding 11,160 test frames. These videos have no ground-truth clean counterparts and are used exclusively for VLM preference evaluation.

\section{Training Details}
\label{sec:appendix_training}

All existing methods are trained from scratch on each dataset following their original training configurations. To ensure a fair comparison, we match the total number of gradient updates to each method's original setting by adjusting the number of epochs according to our dataset size. Tab.~\ref{tab:training_details} summarizes the key hyperparameters for each method.

\paragraph{Our baseline.} We finetune \texttt{Wan 2.2-TI2V-5B} \cite{Wan} using rank-96 LoRA for 1,400 steps on four NVIDIA H200 GPUs. Each training sample is a 1280$\times$720 video of 90 frames, with a batch size of 1 per GPU. We use the AdamW optimizer with a learning rate of 1e-4, a 50-step linear warmup, weight decay of 1e-2, and betas of 0.9 and 0.95. This finetuning completes in approximately 35 hours and consumes roughly one epoch of training data. Unlike existing methods, our baseline operates at full 1280$\times$720 resolution without patch decomposition.

\begin{table*}[t]
\centering
\caption{Training hyperparameters for each method. All existing methods are trained from scratch; our baseline finetunes a pretrained video generation model using LoRA. Input type indicates whether the method processes single images (Img) or video clips of $N$ consecutive frames (Vid-$N$). Patch size is the spatial crop used during training. Total iterations (Iters) are matched to each method's original setting (see Tab.~\ref{tab:iteration_matching}).}
\label{tab:training_details}
\resizebox{\textwidth}{!}{%
\begin{tabular}{llccccccc}
\toprule
 & \textbf{Architecture} & \textbf{Input} & \textbf{Patch} & \textbf{Batch} & \textbf{Iters} & \textbf{Optimizer} & \textbf{LR} & \textbf{Loss} \\ \midrule
\textbf{ESTINet \cite{ESTINet}} & ResNet18 + R-CLSTM + BiLSTM & Vid-5 & 224 & 5 & 79.7K & Adam & 1e-4 & MSE \\
\textbf{RDD-Net \cite{RDDNet}} & RDD-Net & Vid-7 & 128 & 8 & 320K & Adam & 1e-4 & L1 \\
\textbf{RLP \cite{RLP}} & Uformer-T + RLP/RPIM & Img & 256 & 20 & 74.7K & AdamW & 2e-4 & Charbonnier \\
\textbf{Turtle \cite{Turtle}} & Transformer & Vid-5 & 128 & 1 & 200K & Adam & 4e-4 & L1 \\
\textbf{UConNet \cite{UConNet}} & UConNet + AngleNet & Img & 128 & 16 & 375K & Adam & 1e-3 & MSE \\
\textbf{WeatherDiff \cite{WeatherDiff}} & DDPM UNet (83M) & Img & 64 & 80$^\dagger$ & 697K & Adam & 2e-5 & $\epsilon$-pred \\
\textbf{NightRain \cite{NightRain}} & 3D DiT (109M) & Vid-4 & 64 & 8$^\dagger$ & 200K & Adam & 2e-5 & $\epsilon$-pred \\ \midrule
\textbf{Our Baseline} & \texttt{Wan2.2-TI2V-5B} (LoRA $r$=96) & Vid-90 & Full & 4 & 1.4K & AdamW & 1e-4 & Flow matching \\ \bottomrule
\multicolumn{9}{l}{\footnotesize $^\dagger$ Effective samples per step (batch size $\times$ patches per sample). NightRain: $2 \times 4 = 8$. WeatherDiff: $5 \times 16 = 80$.}
\end{tabular}%
}
\end{table*}

\paragraph{Sliding window configuration.} Since existing methods operate on fixed-size patches rather than full 1280$\times$720 frames, we apply a non-overlapping sliding window strategy at both training and inference time. Each frame is padded with zeros (at the bottom and right edges) to the nearest multiple of the patch size, then partitioned into non-overlapping tiles. After processing, the tiles are stitched back and cropped to the original 1280$\times$720 resolution. For video models, consecutive frames are grouped into non-overlapping temporal windows matching each method's input length. WeatherDiff uses an overlapping grid with stride $r$=16 instead of non-overlapping patches, following its original inference procedure. Tab.~\ref{tab:sliding_window} lists the patch sizes, padded resolutions, and number of patches per frame for each method.

\begin{table}[t]
\centering
\caption{Sliding window configuration for inference at 1280$\times$720 resolution. Since each method operates on fixed-size patches, the original frame must first be enlarged to fit an exact grid of patches. \textbf{Padded Size} is the resulting resolution after zero-padding the bottom and right edges of the 1280$\times$720 frame to the nearest multiple of the \textbf{Patch Size}; for example, with 224$\times$224 patches, 1280$\times$720 is padded to 1344$\times$896 (= 6$\times$224 by 4$\times$224). The padded frame is then partitioned into a non-overlapping grid of tiles. WeatherDiff uses overlapping patches with stride $r$=16 instead of non-overlapping tiling, following its original inference procedure. Our baseline processes full-resolution frames without patch decomposition.}
\label{tab:sliding_window}
\fontsize{8pt}{10pt}\selectfont
% \resizebox{\textwidth}{!}{%
\begin{tabular}{lcccc}
\toprule
 & \textbf{Patch Size} & \textbf{Padded Size} & \textbf{Grid} & \textbf{Patches/Frame} \\ \midrule
\textbf{ESTINet \cite{ESTINet}} & 224$\times$224 & 1344$\times$896 & 6$\times$4 & 24 \\
\textbf{RDD-Net \cite{RDDNet}} & 128$\times$128 & 1280$\times$768 & 10$\times$6 & 60 \\
\textbf{RLP \cite{RLP}} & 256$\times$256 & 1536$\times$768 & 6$\times$3 & 18 \\
\textbf{Turtle \cite{Turtle}} & 128$\times$128 & 1280$\times$768 & 10$\times$6 & 60 \\
\textbf{UConNet \cite{UConNet}} & 128$\times$128 & 1280$\times$768 & 10$\times$6 & 60 \\
\textbf{WeatherDiff \cite{WeatherDiff}} & 64$\times$64 ($r$=16) & Overlapping & 77$\times$42 & 3,234 \\
\textbf{NightRain \cite{NightRain}} & 64$\times$64 & 1280$\times$768 & 20$\times$12 & 240 \\ \midrule
\textbf{Our Baseline} & Full frame & N/A & 1$\times$1 & 1 \\ \bottomrule
\end{tabular}%
% }
\end{table}

\paragraph{Iteration matching.} Tab.~\ref{tab:iteration_matching} shows how we match training iterations between the original settings and our setup. For each method, we compute the total gradient updates from the original paper (original dataset size $\times$ original epochs / original batch size) and reproduce a comparable number using our dataset by adjusting the number of epochs.

\begin{table*}[t]
\centering
\caption{Iteration matching between original training settings and our setup. For each method, we compute the total gradient updates from the original paper (dataset size $\times$ epochs / batch size) and reproduce a comparable number on our dataset by adjusting the number of epochs. Our baseline is finetuned for only 1,400 steps, consuming approximately one epoch of training data.}
\label{tab:iteration_matching}
\resizebox{\textwidth}{!}{%
\begin{tabular}{lcccccccc}
\toprule
\multirow{2}{*}{} & \multicolumn{4}{c}{\textbf{Original Setting}} & \multicolumn{4}{c}{\textbf{Our Setting (UENR-600K)}} \\ \cmidrule(lr){2-5} \cmidrule(lr){6-9}
 & \textbf{Dataset} & \textbf{Size} & \textbf{Epochs} & \textbf{Iters} & \textbf{Size/Epoch} & \textbf{Batch} & \textbf{Epochs} & \textbf{Iters} \\ \midrule
\textbf{ESTINet \cite{ESTINet}} & NTU-derain \cite{NTURain} & 3,023 & 120 & $\sim$181K & 99,640 & 5 & 4 & $\sim$79.7K \\
\textbf{RDD-Net \cite{RDDNet}} & RainMotion \cite{RDDNet} & $\sim$5,000 & 500 & $\sim$2.5M & 71,171 & 8 & 36 & $\sim$320K \\
\textbf{RLP \cite{RLP}} & GTAV-NightRain \cite{GTAVNightRain} & 5,000 & 250 & $\sim$312K & 498,201 & 20 & 3 & $\sim$74.7K \\
\textbf{Turtle \cite{Turtle}} & (iter-based) & -- & 200K iter & 200K & 99,640 & 1 & (iter) & 200K \\
\textbf{UConNet \cite{UConNet}} & Rain100/800 \cite{JORDER, IDCGAN} & 40,000$^*$ & 150 & $\sim$375K & 40,000$^*$ & 16 & 150 & $\sim$375K \\
\textbf{WeatherDiff \cite{WeatherDiff}} & AllWeather \cite{WeatherDiff} & 1,771 & 1,775 & $\sim$197K & 498,201 & 80$^\dagger$ & 7 & $\sim$697K \\
\textbf{NightRain \cite{NightRain}} & SynNightRain \cite{SynNightRain} & $\sim$1,000 clips & (iter) & 200K & 100,000 & 8$^\dagger$ & (iter) & 200K \\ \midrule
\textbf{Our Baseline} & -- & -- & -- & -- & 498,201 & 4 & $\sim$1 & 1.4K \\ \bottomrule
\multicolumn{9}{l}{\footnotesize $^*$ UConNet subsamples 40,000 images per epoch from the full dataset.} \\
\multicolumn{9}{l}{\footnotesize $^\dagger$ Effective samples per step (see Tab.~\ref{tab:training_details}). WeatherDiff iterations are matched by total patch count rather than gradient steps.}
\end{tabular}%
}
\end{table*}

\section{Deraining Quality Evaluation Using VLM as a Judge}
\label{sec:appendix_vlm}

We use \texttt{claude-sonnet-4-6} \cite{Claude} as the VLM judge for deraining quality evaluation on the 124 real nighttime rain videos from Pexels \cite{Pexels} (90 frames each). The VLM receives sampled frames from competing methods and selects the best result. To ensure reproducibility, we provide the complete evaluation prompt below.

\paragraph{Evaluation prompt.} For each comparison, the VLM receives the rainy input image and $N$ candidate derained images labeled with randomized anonymous letters (A, B, ...). The prompt instructs:

\begin{lstlisting}[style=promptbox]
You are conducting a blind evaluation of image deraining quality. For each comparison, you will view a rainy input image and N candidate derained images labeled with letters. Your job is to pick the candidate with the best deraining quality.

Compare candidates on the following criteria:
(1) Rain removal: how effectively are rain streaks, rain drops, rain curtains, and rain-induced fog removed?
(2) Detail preservation: are scene structures, textures, edges, and fine details maintained?
(3) Artifact avoidance: does the result avoid introducing blocking artifacts, color distortions, excessive blurring, or visual hallucinations?
(4) Overall quality: considering all factors, which result looks most natural and visually clean?

For each comparison, view the input image first for context, then view all candidates, choose the best one based on the criteria above, and provide a one-sentence justification.
\end{lstlisting}

The VLM must provide a written justification before making its selection to encourage careful analysis. Method labels are randomized per evaluation item using a fixed seed to mitigate position bias.

\begin{figure}[t]
    \centering
    \includegraphics[width=\linewidth]{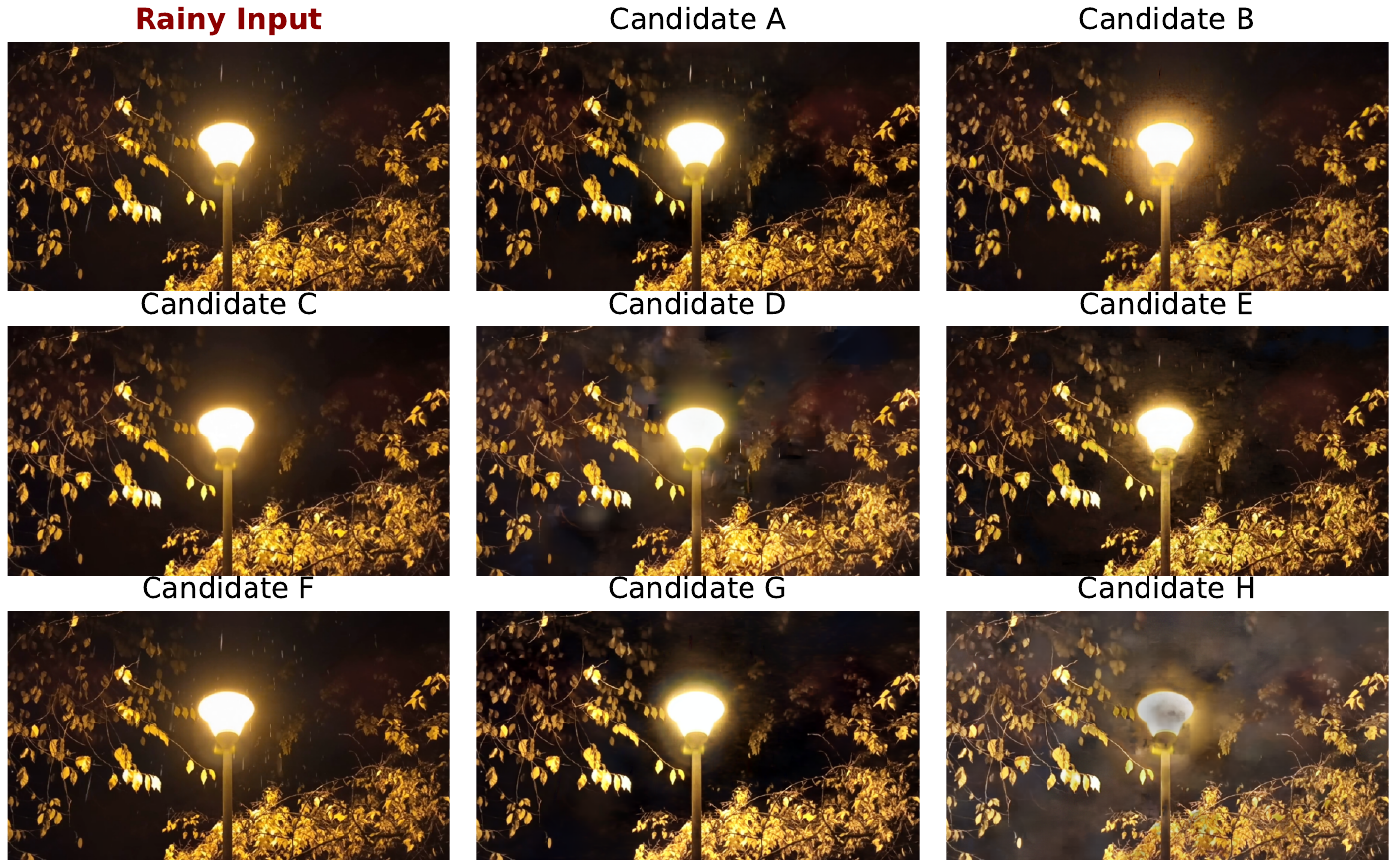}
    \caption{Example of a multi-way VLM evaluation item. The VLM receives the rainy input (top-left) and eight candidate derained images labeled A through H. Method-to-label assignments are randomized per evaluation item, so the VLM cannot learn positional patterns. For this example, the VLM selected Candidate C with the justification: \textit{``Candidate C produces the cleanest rain removal with natural color and detail preservation, free of the haze artifacts seen in H and the residual streaks in A and F.''}}
    \label{fig:vlm_deraining_example}
\end{figure}

\paragraph{Multi-way comparison.} The VLM views results from all eight methods for the same frame and selects the single best output. We evaluate one frame per video across all 124 videos, yielding 124 comparisons per experiment (covering 1.1\% of the 11,160 total test frames). Each comparison presents 9 images to the VLM: 1 rainy input and 8 candidate outputs.

\paragraph{Pairwise comparison.} For comparing models trained on different datasets (e.g., our dataset versus SynNightRain), the VLM views two results side by side for three frames per video across all 124 videos, yielding 372 comparisons per experiment (covering 3.3\% of test frames). A tie option is available when the two results are visually indistinguishable. For pairwise evaluations, the preference rate is computed as $\text{wins} / (\text{wins} + \text{losses})$, excluding ties from the denominator.

\paragraph{Parallelization and batching.} Evaluations are distributed across multiple independent VLM instances (up to 4 concurrent), each processing a disjoint batch of evaluation items. Each instance is self-contained and receives the full prompt with its assigned images. This parallelization does not affect evaluation quality since each item is evaluated independently.

\section{Temporal Consistency Evaluation Using VLM as a Judge}
\label{sec:appendix_temporal}

We use \texttt{claude-opus-4-6} \cite{Claude} as the VLM judge for temporal consistency evaluation, rating each method's output independently on a 1 to 5 scale. Unlike the pairwise deraining evaluation above, which compares methods against each other, the temporal evaluation rates each (model, video clip) pair independently to avoid conflating spatial quality with temporal smoothness.

\paragraph{Frame selection and difference maps.} For each video clip, we select 4 consecutive frames from the temporal midpoint (e.g., frames 44 to 47 for 90-frame clips). In addition to raw frames, we pre-compute 3 inter-frame difference maps by taking the per-pixel absolute difference between consecutive output frames, converting to grayscale, and amplifying by a factor of 10. In a temporally consistent result, only regions with real scene motion appear bright in the difference maps; flickering artifacts appear as bright patches in regions that should be static (e.g., building facades, sky, parked cars).

\begin{figure}[t]
    \centering
    \includegraphics[width=\linewidth]{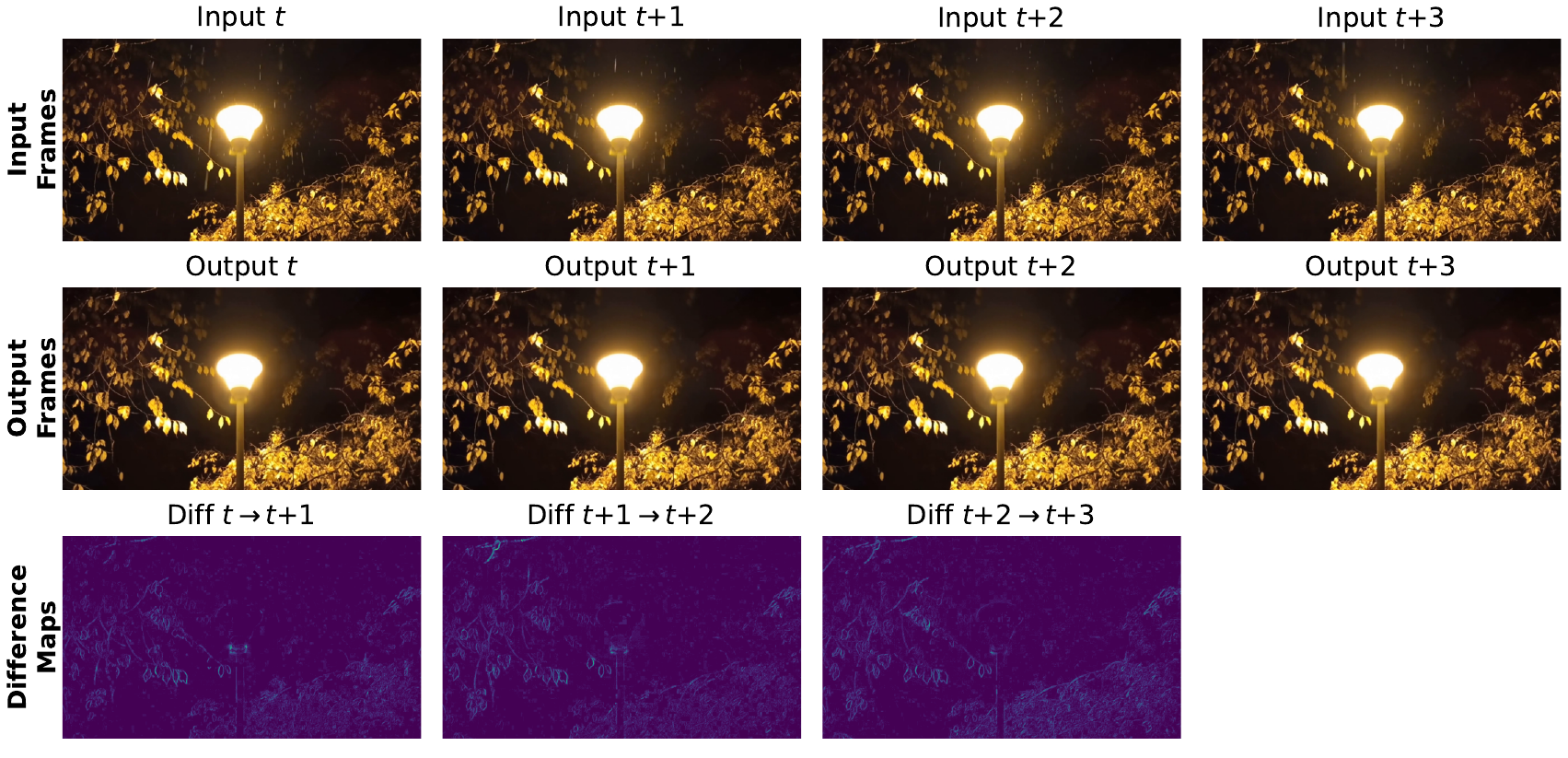}
    \caption{Example of a temporal consistency VLM evaluation item. \textbf{Top:} four consecutive rainy input frames providing scene context. \textbf{Middle:} four consecutive derained output frames to evaluate. \textbf{Bottom:} three inter-frame difference maps (amplified 10$\times$), where bright regions indicate large pixel changes between adjacent output frames. In a temporally consistent result, only regions with real scene motion (e.g., swaying leaves) appear bright, while static regions (e.g., lamp post, sky) remain dark. Flickering artifacts would manifest as bright patches in static areas.}
    \label{fig:vlm_temporal_example}
\end{figure}

\paragraph{Evaluation prompt.} Each evaluation item presents the VLM with 11 images: 4 rainy input frames (for scene context), 4 derained output frames (to evaluate), and 3 inter-frame difference maps. The prompt instructs:

\begin{lstlisting}[style=promptbox]
You are evaluating the temporal consistency of video deraining results. Rate the following video sequence on a 1 to 5 scale:

5 (Excellent): No visible temporal artifacts. Smooth, natural frame-to-frame transitions.
4 (Good): Very minor temporal inconsistencies, barely noticeable.
3 (Fair): Noticeable temporal artifacts. Some flickering, inconsistent deraining across frames, or minor ghosting.
2 (Poor): Significant temporal artifacts. Obvious flickering, prominent frame-to-frame inconsistencies.
1 (Very Poor): Severe temporal artifacts. Extreme flickering, frames appear unrelated.

What to look for: flickering (unexpected brightness or color changes in static regions), inconsistent rain removal across consecutive frames, ghosting or bleeding artifacts, and whether real scene motion appears smooth and continuous. Use the difference maps to detect subtle flickering: bright patches in regions that should be static indicate temporal artifacts.

Read all images (input frames, output frames, difference maps), then rate 1 to 5 and give a one-sentence reason.
\end{lstlisting}

\paragraph{Batching and scale.} Items are grouped into batches of 4 (44 images per batch) and shuffled with a fixed seed so that consecutive batches contain a mix of different models. Up to 4 VLM instances process batches in parallel. We evaluate all 124 clips per model across both training datasets (8 methods $\times$ 2 datasets = 16 configurations), yielding 1,984 total evaluations. Per-model scores are aggregated as the mean rating across all evaluated clips.

\paragraph{Evaluation cost.} Tab.~\ref{tab:vlm_cost} reports the API cost for reproducing all VLM evaluations in this paper. Each 1280$\times$720 image consumes 1,229 input tokens, and the 5,208 evaluations use 41.7M input tokens and 0.7M output tokens in total. The temporal consistency evaluation accounts for 63\% of the total cost because each evaluation item presents 11 images to the VLM.

\begin{table}[t]
\centering
\caption{API cost for reproducing all VLM evaluations using \texttt{claude-opus-4-6} \cite{Claude}. Each 1280$\times$720 image consumes 1{,}229 input tokens; the 5{,}208 evaluations use 41.7M input tokens and 0.7M output tokens in total. The temporal evaluation is the most expensive because each item requires 11 images (4 input frames, 4 output frames, and 3 difference maps).}
\label{tab:vlm_cost}
\fontsize{8pt}{10pt}\selectfont
\begin{tabular}{lccr}
\toprule
 & \textbf{Input (MTok)} & \textbf{Output (MTok)} & \multicolumn{1}{c}{\textbf{Cost (\$)}} \\ \midrule
Pairwise quality (Tab.~\ref{tab:vlm_eval_per_method}) & 11.57 & 0.45 & 69.00 \\
Multi-way quality (Tab.~\ref{tab:vlm_eval_per_dataset}) & 2.82 & 0.04 & 15.00 \\
Temporal consistency (Tab.~\ref{tab:temporal_consistency}) & 27.32 & 0.20 & 141.55 \\ \midrule
\textbf{Total} & \textbf{41.71} & \textbf{0.69} & \textbf{225.55} \\ \bottomrule
\end{tabular}
% \begin{tabular}{lrrrr}
% \toprule
%  & \textbf{Evals} & \textbf{Img./Eval} & \textbf{Total Img.} & \textbf{Cost (\$)} \\ \midrule
% Pairwise quality (Tab.~\ref{tab:vlm_eval_per_method}) & 2{,}976 & 3 & 8{,}928 & 69.00 \\
% Multi-way quality (Tab.~\ref{tab:vlm_eval_per_dataset}) & 248 & 9 & 2{,}232 & 15.00 \\
% Temporal consistency (Tab.~\ref{tab:temporal_consistency}) & 1{,}984 & 11 & 21{,}824 & 141.55 \\ \midrule
% \textbf{Total} & \textbf{5{,}208} & & \textbf{32{,}984} & \textbf{225.55} \\ \bottomrule
% \end{tabular}
\end{table}

\section{Inference Time}
\label{sec:appendix_inference}

Tab.~\ref{tab:inference_time} reports the per-frame and total inference time for each method on the Ours test set (1,800 frames at 1280$\times$720). All existing methods are measured on a single NVIDIA A5000 GPU. Non-diffusion methods complete within 24 minutes (UConNet) to 185 minutes (Turtle). Diffusion-based methods require significantly longer due to iterative denoising over many patches: WeatherDiff takes 5.50 hours and NightRain takes 37.41 hours. Our baseline processes full 1280$\times$720 frames without patch decomposition, completing 1,800 frames in 12.31 hours on a single NVIDIA H200 GPU; based on FP32 throughput scaling, the estimated A5000-equivalent time is approximately 29.5 hours.

\begin{table}[t]
\centering
\caption{Inference time on the Ours test set (1,800 frames at 1280$\times$720) using a single GPU. Non-diffusion methods complete within minutes to hours, while diffusion-based methods (NightRain, WeatherDiff) require significantly longer due to iterative denoising over many patches. Our baseline processes full-resolution frames without patch decomposition.}
\label{tab:inference_time}
\fontsize{8pt}{10pt}\selectfont
\begin{tabular}{lcc}
\toprule
 & \textbf{s/frame} & \textbf{Total Time} \\ \midrule
\textbf{ESTINet \cite{ESTINet}} & 4.34 & 130.2 min \\
\textbf{RDD-Net \cite{RDDNet}} & 2.79 & 83.7 min \\
\textbf{RLP \cite{RLP}} & 1.09 & 32.7 min \\
\textbf{Turtle \cite{Turtle}} & 6.18 & 185.4 min \\
\textbf{UConNet \cite{UConNet}} & 0.81 & 24.3 min \\
\textbf{WeatherDiff \cite{WeatherDiff}} & 11.00 & 5.50 h \\
\textbf{NightRain \cite{NightRain}} & 74.81 & 37.41 h \\
\textbf{Our Baseline} & 24.61 & 12.31 h$^\dagger$ \\ \bottomrule
\multicolumn{3}{l}{\footnotesize $^\dagger$ Measured on NVIDIA H200; all others on A5000.} \\
\multicolumn{3}{l}{\footnotesize \phantom{$^\dagger$} Scaled by FP32 throughput (2.4$\times$), est.\ $\sim$29.5 h on A5000.}
\end{tabular}
\end{table}

\end{document}